# weighted CapsuleNet networks for Persian multi-domain sentiment analysis

*Mahboobeh Sadat Kobari[a], Nima Karimi[b], Benyamin Pourhosseini[c], Ramin Mousa[d]*

[a]*Islamic Azad University, Tehran, Iran, Email:* Mahboobe.kobari@gmail.com
[b]*University of Yazd, Yazd, Iran, Email:* Nimakarimi@gmail.com, Nimakarimi@Stu.yazd.ac.ir
[c] *K.N. Toosi University of technology Tehran, Tehran, Iran, Email:* Pourhosseini@email.kntu.ac.ir
[b]*University of Zanjan, Zanjan, Iran, Email:* Raminmousa@znu.ac.ir

ABSTRACT

Sentiment classification is a fundamental task in natural language processing, assigning one of the three classes, positive, negative, or neutral, to free texts. However, sentiment classification models are highly domain dependent; the classifier may perform classification with reasonable accuracy in one domain but not in another due to the Semantic multiplicity of words getting poor accuracy. This article presents a new Persian/Arabic multi-domain sentiment analysis method using the cumulative weighted capsule networks approach. Weighted capsule ensemble consists of training separate capsule networks for each domain and a weighting measure called domain belonging degree (DBD). This criterion consists of TF and IDF, which calculates the dependency of each document for each domain separately; this value is multiplied by the possible output that each capsule creates. In the end, the sum of these multiplications is the title of the final output, and is used to determine the polarity. And the most dependent domain is considered the final output for each domain. The proposed method was evaluated using the Digikala dataset and obtained acceptable accuracy compared to the existing approaches. It achieved an accuracy of 0.89 on detecting the domain of belonging and 0.99 on detecting the polarity. Also, for the problem of dealing with unbalanced classes, a cost-sensitive function was used. This function was able to achieve 0.0162 improvements in accuracy for sentiment classification. This approach on Amazon Arabic data can achieve 0.9695 accuracies in domain classification.

## 1- Introduction:

Sentiment classification is one of the most basic tasks in natural language processing. In recent decades, many supervised machine learning methods, such as Naive Bayes, support vector machines, and neural networks, have been used for this task (۱, ۲). However, sentiment classification models are highly domain-dependent, leading to the demand for a large amount of training data for each domain. The reason is that there are usually different words and phrases in other domains, and even one word in various fields may reflect different emotional poles. For example, the word easy is often used when the sentence conveys positive feelings in the area of the child's products (e.g., it is easy for him to hold…). But in the realm of film criticism, easy can sometimes convey negative sentiments (e.g., it's easy to guess the ending of this movie).Therefore, using the resources available in all domains is valuable to improve emotion classification performance in some specific domains.

This article presents an approach called WcapsuleE for the multi-domain sentiment analysis of the Persian/Arabic language, which uses cumulative capsule networks to calculate the polarity and domain of belonging. The critical feature of this approach is that with the help of a domain membership calculation criterion, the membership domain and polarity are detected simultaneously. The results of the tests by these approaches on the Digikala data set (including ten different domains) show that the proposed approach performs better in all domains than the existing methods.

**2- Related Work:**

Sentiment analysis has been studied in various application domains (۳-۵). Figure 2 shows the percentage of algorithms proposed in this field in the Scopus database.

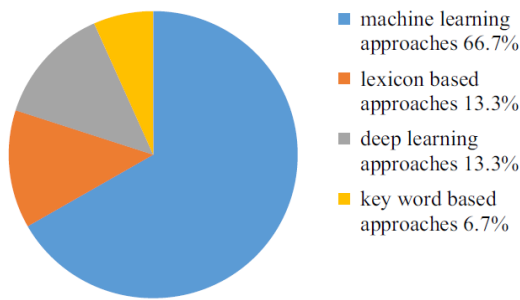

**Figure1.** The percentage of using different algorithms and approaches for sentiment analysis.

**2.1 Approaches based on deep learning**:

Deep learning includes different types of neural networks such as ANN[1], CNN[2], RNN[3], LSTM[4], GRU[5], and CapsuleNet, and below, we will briefly discuss the work done in the field of sentiment analysis using these networks.

1- **Convolutional Neural Networks (CNN):** This network actually eliminates the fully connected connections in artificial neural networks. These networks are a kind of forward neural networks with the following properties (۶):

   a: The convolution Layer, b: Sparse Connection, c: Parameter Sharing, d: Pooling

2- **Recurrent neural networks:** Recurrent neural networks are forward-facing networks that add the concept of time to the model, defined through edges in adjacent steps. In these networks, two problems vanishing gradient and

---
[1] Artificial Neural Network
[2] Convolution Neural Network
[3] Recurrent Neural Network
[4] Long-Short Term Memory
[5] Gated Recurrent Unit

۲

Exploding gradient, occur during the backpropagation in long time steps. LSTM networks have been proposed as a solution for gradient fading by Schmidhuber (7). Another type of neural network offered to solve the gradient fading problem is GRU networks, which were first proposed about machine translation(8). Each GRU node has two updates and resets gates. The update gate decides which part to update, and the reset gate allows previous computational states to be forgotten.

3- **Capsule neural networks:** Another limitation of CNN is that its neurons are activated based on the probability of certain features (not all features), so they are not trained on the relationships of these features. Capsule networks (9) have been presented to solve these problems. Capsule means to cover and protect, which is used because, by this method, all the critical features of the input data are preserved. The vital difference between capsule networks and convolution is that, unlike convolution, where values are stored numerically (scalar), they are stored in the capsule as vectors.

**2.2 multi-domain problem:**

In multi-domain sentiment analysis, as mentioned, the goal is to train a classification on a set of domains to solve the problem of dependence on a specific domain (10). The main idea of the stated approachis to convert the raw input texts into an embedded representation based on word2vec and use it to predict the polarity of opinions and to identify the domain of affiliation in parallel with the polarity discovery process. This approach used LSTM memory cells to create a deep network. In this article, the output value of the beam layer is used to determine the range of belonging as follows:

$$y = softmax(W_y c + b_y)$$

where $W_y \in R^{h*|D|}$ . $b_y \in R^{|D|}$ and |D| is the number of domains in the training set. Also, the value of the beam layer has been used to determine the polarity of each domain as follows:

$$Z = tanh(W_z c + b_z)$$

Some examples of the most important research done in Farsi SA for Digikala are given in Table 1. All these investigations work on single-domain SA.

| Model | Ref. | Domain Type | Evaluation criteria |
|---|---|---|---|
| Different classification models and different feature selection approaches | (11) | Single domain | Precision=91.22% Recall=91.71% F1=91.46% |

| Unsupervised models and neural network | (12) | Single domain | Precision=73.7% Recall=99.1% F1=58.6% |
| Conceptual dictionary of words and polarity recognition | (13) | Single domain | Accuracy=86% F1=80% Recall =75% |
| Dictionary-oriented approach | (14) | Single domain | Accuracy=94% F1=89% Recall =88% Precision=90% |
| Polarity detection and support vector machine | (15) | Single domain | F1=90.15% Precision=93.03% Recall =87.42% |

**Table 1:** Single domain sentiment analysis on Digikala data

### 2.3 Single-domain and multi-domain Arabic sentiment analysis:

Even though Arabic is one of the most common languages in the world, it has received little attention in sentiment analysis.

Most of the SA studies have worked on natural languages such as English, Chinese, and Arabic. NLP in Arabic is still in its early stages (۱۶). It lacks advanced resources and tools. Therefore, the Arabic language still faces challenges in NLP tasks due to the complexity of its structure, history, and different cultures (۱۷).

Many tools and approaches, either semantic approaches or machine learning (ML) approaches, have been used in the literature to perform the SA task. Most of them are designed to administer SA in English, a scientific language. The semantic approach extracts emotional words and calculates their polarities based on emotional vocabulary. In contrast, to build a new model, ML classifiers train the annotation data after transforming it into feature vectors to infer the specific features used in a particular class. Finally, the new model can be used to predict new data classes. It is worth noting that these approaches can be adapted to other languages, such as Arabic. Compared to other languages, the Arabic language has made less effort; However, hundreds of studies have been proposed for ASA. Since its introduction a decade ago, ASA has become one of the most popular forms of information extraction from surveys. The table below summarizes examples of these approaches for the Arabic language.

| Model | Reference | Domain Type | Model type | Performance |
|---|---|---|---|---|
| GRU | (18) | Single Domain | Deep learning | Accuracy= 83.98% |
| SVM and Naive Bayes | (19) | Single Domain | Machine learning | SVM Accuracy=91.40%<br><br>Naïve bayes Accuracy=88.08% |
| TFICF | (20) | Multi-Domain | Lexicon Based | Accuracy= 89% |
| Arabic Ontology-Based | (21) | Multi-Domain | Ontology-Based | Accuracy= 79.20% |

**Table 2: Single-domain and multi-domain sentiment analysis on Arabic data**

### 3- The proposed approach of WcapsuleE:

Most neural networks transform input vectors into output vectors by combining matrix multiplications and nonlinear functions. The nonlinearities in neural networks perform the same operation independently on each vector element. Usually, intermediate representation elements are referred to as the activation of a neuron that follows the neural pattern in the brain. Capsule networks differ from regular neural networks in some critical ways. In capsule networks, a predefined set of neurons in an arbitrary layer is called a capsule; more importantly, there should be more than one capsule; that is, the set should be smaller than the entire set of neurons in the layer. The combined activation of neurons in a capsule is called a state. Unlike normal networks, the capsule network sends features to the next layer in each layering, and in normal networks, these features are in the form of scalar values. A capsule is a group of neurons whose activity vector represents the instantiation parameters of a specific entity type, such as an object or a piece of an object (22).

The capsule idea was first described by Hinton et al. in Transforming Autoencoders (22). In this paper, they explain how convolutional neural networks can recognize objects but not their position in space. This is due to pooling layers, which can eliminate the distance between features. Instead of aiming for viewpoint invariance, ANN models should aim for viewpoint equivalence. This can be done by using capsules, representing a single entity in the image, and a vector of instance parameters representing the entity's characteristics, along with the probability that this entity exists in its bounded domain shows (23).

(23) defined a capsule as a group of neurons with sampling parameters represented by activity vectors, where the length of the vector represents the probability of feature presence. This network consists of the convolutional, primary capsule (PC), and class capsule layers. The initial capsule layer is the first capsule layer, followed by an unspecified number of capsule layers until the last capsule layer, also called the class capsule layer. The convolution layer does the feature extraction from the image, and the output is entered into the initial capsule layer. Each capsule i (where $1 \leq i \leq N$) in layer $l$ has an activity vector $u_i \in R$ to encode spatial information in the form of sample parameters. The output vector $u_i$ of the lower level capsule, i is input to all capsules in the next layer l+1. The jth capsule in layer l+1 receives $u_i$ and finds its product with the corresponding weight matrix $W_{ij}$. The resulting vector $\hat{u}_{j|i}$ is capsule $i$ in the level l transformation of the entity represented by capsule j at level $l+1$. The prediction vector of a PC, $\hat{u}_{j|i}$, shows how closely the initial capsule i is related to the class capsule j.

There are different approaches to capsule networks for textual data. For example, this network has been used for sentiment analysis (۲۴–۲۶), text classification (۲۷–۲۹), etc. The noteworthy point is that these networks require CNN or RNN networks (LSTM, GRU, IndRNN) to create initial vectors. In the proposed approach, we use Bi-GRU networks to generate initial vectors. As mentioned, another type of RNN network is GRU networks, which are proposed to solve gradient loss. These networks have two gates, rest and update, faster than LSTM and can learn more extended time series.

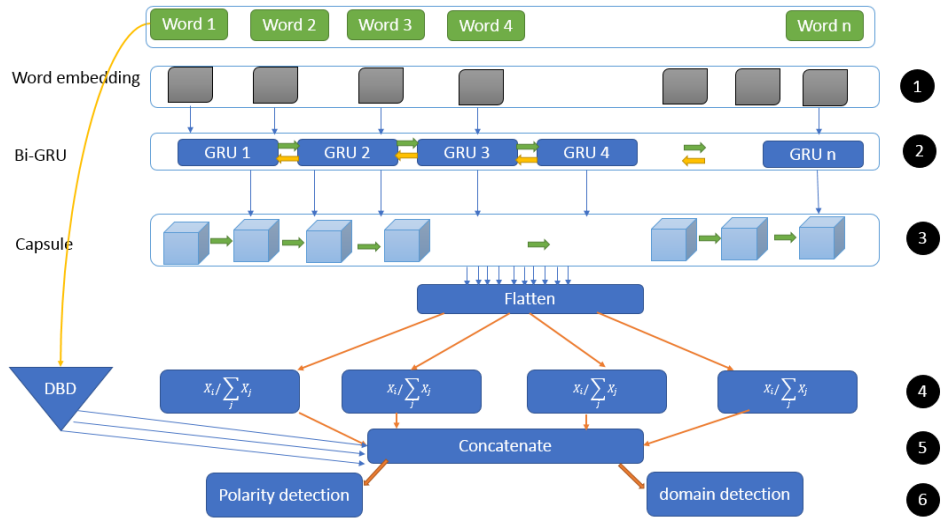

**Figure 1:** Propused model flowchart.

In addition, as mentioned, in capsule networks, each neuron has probability and properties related to features, which can be modeled as a vector of features. We can detect any inconsistencies in the inputs using the information stored in these vectors. The proposed approach considers the combination of capsule and GRU networks. This approach has been used in work (۳۰) to classify emotion as domain-dependent or single-domain. In our proposed approach, the two-way mode of the GRU network (Bi-GRU) and its combination with the capsule network are used for multi-domain sentiment analysis. The capsuleNet method for single-domain sentiment analysis has been used previously in (۳۰). Our proposed approach is very similar to (۳۱) with the difference that we try to train the network for each domain separately and combine it with DBD to get the final polarity. The proposed approach includes six basic steps as follows:

1. **Words embedding:** We considered each document a sequence of words, then removed all stop-words and punctuation marks in this sequence. Eliminating these items affects the overall accuracy and makes the classifier learn better features. We also removed all the rare words repeated only once in the general words. To better understand our topic, we can create a document as a sequence of words as $[x^{<1>}, x^{<2>}, x^{<3>}, ..., x^{<n>}]$, where $x^{<1>} \in R^k$ and $k$ refer to the dimensions of the i[th] word vector of this sequence. One of the limitations of CNN networks is that the dimensions of the inputs must be equal. On the other hand, the length of the sequence of input entries is different. For this purpose, we considered the maximum length of the document in all the data as the threshold and padded all the documents with a value of 0. This work is called "Zero padding". Therefore, the result of this padding can be displayed as follows:

$$S = [x^{<1>} \oplus x^{<2>} \oplus x^{<3>} \oplus ... \oplus x^{<n>}]$$

Where $\oplus$ is the concatenation operator. Then, for each of these words in sequence S, we extracted the Fasttext embedding vectors. Each document in this layer is converted into dense vectors according to the pattern mentioned in the previous section. These dense vectors are obtained by Fasttext Words embedding. For this, we used the pre-trained Fasttext dataset with 400 dimensions. The output of this layer for each document is equal to the following value:

$$out_{embed} = M * E$$

Where M is equal to the maximum length of the document in the entire data set, and E is the size of the embedded vectors. Fasttext with size 400 is used here. The figure below shows an overview of the creation of initial vectors for two different views:

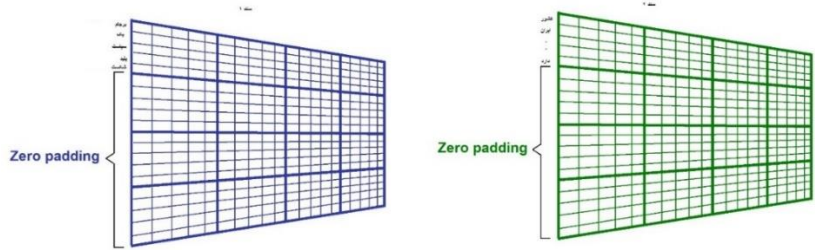

**Figure 3: Zero-padding operation to create equal comments for neural networks**

2. **Bi-GRU layer:** The input of the Bi-GRU layer is the embedded vectors taken from the embedding layer. Suppose we display these vectors as $Out_{embed} = [X_1, X_2, ..., X_n]$. The input of Bi-GRU in step T is the vector $X \in R^{400}$. Then $h_t = h_1, h_2, ..., h_t$ which show the sequence of hidden vectors in Bi-GRU and are calculated through the following relationship.

$$z_t = \sigma(w_z x_t + U_z h_{t-1} + b_z)$$
$$r_t = \sigma(w_r x_t + U_r h_{t-1} + b_r)$$
$$h'_t = \sigma(w_h x_t + U_n(r_t \odot h_{t-1}) + b_h)$$
$$h_t = (1 - z)h_{t-1} + z_t h'_t$$

where $z_t$ is the update gate, $z_t$ is the rest gate, $h'_t$ is the conditional gate and $h_t$ is the active output. $w_z, w_r, W_n, U_z, U_r, U_n$ are learnable matrices, $\boldsymbol{b_n}, \boldsymbol{b_z}, \boldsymbol{b_r}$ are learnable biases, σ is the activation function and $\odot$ is the dot multiplication sign between the elements. Bi-GRU networks usually model data in one direction. In some tasks, input sequence inversion can improve network performance. Bidirectional GRU networks (Bi-GRU) (32) process data in both Forward and Backward directions. If $\overrightarrow{h_t}$ is the Forward output for the sequence $x_1^t$ when t=1,2,3,...,t and $\overleftarrow{h_t}$ is the Backward output for $x_1^t$ when $t = t, \ldots .3,2,1$ is, then the Bi-GRU output is obtained through the step-by-step combination of Forward and Backward outputs as $h_t = (\overrightarrow{h_t}.\overleftarrow{h_t})$. This network has twice as many free parameters compared to the unidirectional mode.

3. **Encapsulation layer:** Encoded features are fed into an encapsulation network by the Bi-GRU layer. The encapsulation layer converts the scalar features extracted by the GRU layer into vector-valued encapsulations. If the output of Bi-GRU is equal to $h_i$ and W is the weight matrix, then $v'_{(i|j)}$ which represents the prediction vector obtained from the following equation:

$$v'_{(i|j)} = w_{ij} h_i$$

The set of inputs to a capsule $s_j$ is a weighted set of all prediction vectors $\hat{v}_{(i|j)}$ which is calculated through the following equation:

$$s_j = \sum c_{i,j} . v'_{(i|j)}$$

In this relationship, $c_{i,j}$ is the correlation coefficient, whose value is set by the dynamic routing algorithm. Finally, the Squash function is used as a non-linear function to map the $S_j$ values to the [0-1] range: $v_j = \frac{||z_j||^2}{1+||z_j||^2} . \frac{z_j}{||z_j||}$.

4. **Classification layer:** The output of the proposed Bi-GRUCapsule network differs from conventional approaches because this layer is supposed to be used for two different tasks (polarity detection and domain belonging detection) simultaneously. Initially, the capsule output is flattened into a fully connected layer with ten neurons.

$$P = W_{dense*F}$$

The output *P* should be such that it represents the probability of each of the ten classes. For this purpose, the Softmax function was used, which calculates the value of Pi for each $f_i \in F$ as follows:

$$p_i = \frac{e^{-f_i}}{\sum_{f_j \in F} e^{-f_j}}$$

One of the most critical steps in the adequate performance of the WCapsuleNetE-based method is the appropriate calculation of the parameter belonging to the domain or DBD. The main idea of DBD used in the proposed approach is the use of the TF-IDF concept [44] and taken from the field of information retrieval.

The main goal of TF-IDF is to extract the relationship between a word and a specific domain (based on all documents in the domain) based on two concepts in IR. These two concepts are TF which calculates the occurrence rate of a word in a domain, and IDF, which calculates the repetition rate of the word in each document. The value of TF for the word T in the domain $d_i$ is obtained through the following relation

$$TF(T, d_i) = \frac{n_{T_i}}{N_i}$$

where $n_{T_i}$ is the number of occurrences of word T in domain $d_i$ and $N_i$ is the number of occurrences of all words in domain $d_i$. According to these values, the IDF for the word T can be calculated as follows:

$$IDF(T, d_i) = \frac{n_{T_i}}{\sum_{j=1}^{M} n_{T_i}}$$

where M is the total number of domains in the data set (in our training set, this value is equal to 10), and $n_{T_i}$ is the number of occurrences of the word T in domain i. The value of DBD is obtained through the following relationship based on TF and IDF:

$$DBD(T, d\_i) = TF(T, d_i).IDF(T, d_i)$$

It is enough to detect the domain of belonging to find only the index of the largest value in 10 bits of DBD from this 10-bit vector. If D equals these 10 bits, we will have the following relationship:

$$Domain\ identicication = Arg_i \max\{D_i\}$$

But to detect the polarity, you must first obtain the maximum probability value for each class in Pos and Neg mode, and then create the polarity vector, if this vector is equal to C, then each element of it is obtained through the following relationship.

$$c_i = \begin{cases} pos_i & Pos_i \geq Neg_i \\ -Neg_i & Pos_i < Neg_i \end{cases}$$

Where i refers to the desired domain. In the end, the final polarity for a document d is obtained by multiplying the two vectors C and D. If this larger product equals 0, the polarity is positive; otherwise, it is negative.

$$Polarity(d) = D_d.C_d$$

For better understanding, suppose the following values are obtained for 3 out of 10 domains for document d.

$$D_1(d) = 0.03$$
$$D_2(d) = 0.75$$
$$D_3(d) = 0.40$$
$$C_1(d) = 0.06$$
$$C_2(d) = 0.08$$
$$C_3(d) = -0.03$$

Then, according to the relationship $Arg_i \max\{D_1, D_2, D_3\}$, the domain D2 is desired, and the polarity is also positive for document d.

$$Polarity(d) = 0.03 * 0.06 + 0.75 * 0.08 + 0.40 * -0.03 = 0.049$$

One of the most critical problems of the proposed approach is the lack of control of unbalanced classes in a data set; hence the proposed model may achieve much less accuracy on minority classes. For this purpose, a cost-sensitive function has been used. Cost-sensitive classifiers' main advantage is distinguishing samples into majority and minority classes. According to(33) , misclassification costs can be considered as a clutter matrix, where 0 are negative classes (majority), and 1 are positive classes (minority). Rows are actual classes, and columns are predicted classes. With these conditions, the geometric value and the degree of accuracy can be extracted from the confusion matrix as follows:

$$G_{mean} = \sqrt{\frac{TP}{TP+FN} * \frac{FP}{TN+FP}}$$

$$ACC = \frac{TP+TN}{TP+TN+FP+FN}$$

However, the fixed cost matrix cannot accommodate the unbalanced distribution of local areas, such as the small training sets of Bi-GRUCapsule. Therefore, the proposed method uses a dynamically changing misclassification cost weight. Dynamic weighting can be adaptively updated. We define a cost-sensitive learning strategy to deal with the imbalanced class problem. According to the definition of (34), the overall loss function and optimization is shown in the following equation.

$$E(\theta) = \frac{1}{n^{pos}}\sum_{i=1}^{n^{pos}} LOSS^{pos}(\theta^{pos}, \lambda_n^{pos}) + \frac{1}{n^{neg}}\sum_{i=1}^{n^{neg}} LOSS^{neg}(\theta^{neg}, \lambda_n^{neg})$$

$$\lambda_n = \begin{cases} IR^{overall} * \exp\left(-\frac{G_{mean}^{batch}}{2}\right) * \exp\left(-\frac{ACC^{batch}}{2}\right), & if\ n \in neg) \\ 1, & if\ n \in pos \end{cases}$$

$$\theta^* = argminE(\theta)$$

where $IR^{overall}$ is the overall imbalance ratio. Batches $G_{mean}^{batch}$ and $ACC^{batch}$ are the geometric value and accuracy of the current minibatch training samples, respectively. This trick will be applied with the aim of preventing the absence of minority samples in each minibatch and improving the generalization of the classifiers. In this layer, Softmax is also used to calculate the loss function.

### 4- Data collection:

One of the most critical challenges in the Persian/Arabic language is the lack of a multi-domain SA evaluation protocol. For this purpose, we collected opinions on the DigiKala website to create an evaluation protocol. In this regard, we collected 50,799 opinions in 10 different areas: shoes, perfumes, phones, creams, printers, clothes, books, beds, cars, and gold. The labeling process was completely manual. In choosing the polarity of comments, we considered comments with a positive score of 4 or higher and those with less than two negative. It should be mentioned that the collected data is highly unbalanced, and the number of negative samples is much less than positive samples. Figure 5-16 shows the percentage of data frequency per label, data frequency per domain, and label per domain.

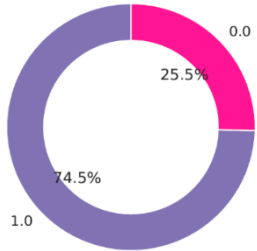

Figure 2: Shoes domain labels frequency.

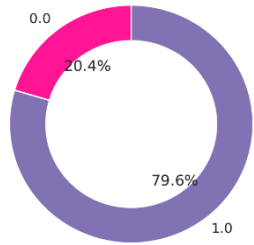

Figure 3: Perfume domain labels frequency.

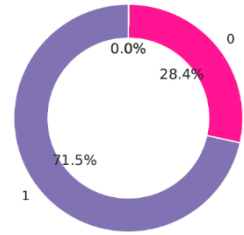

Figure 4 Phone domain labels frequency.

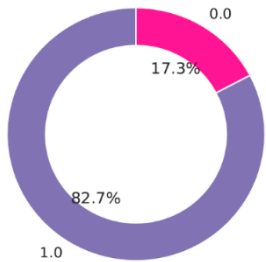

Figure 5: Cold cream domain labels frequency.

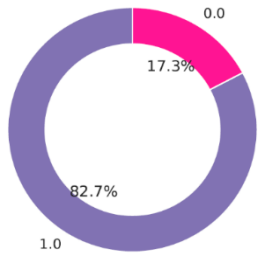

Figure 6: printer domain labels frequency.

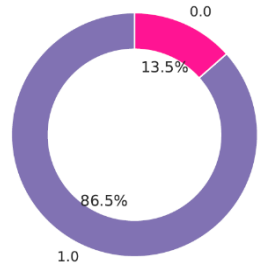

Figure 7: dress domain labels frequency.

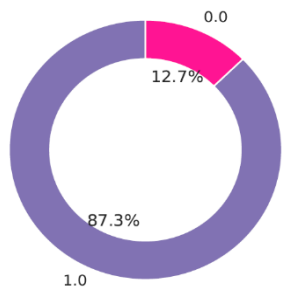

Figure 8: Book domain labels frequency.

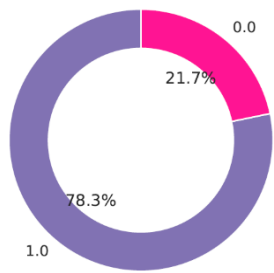

Figure 9: bed domain labels frequency.

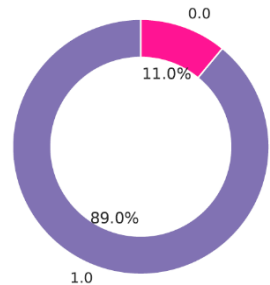

Figure 10 shaving machine domain labels frequency.

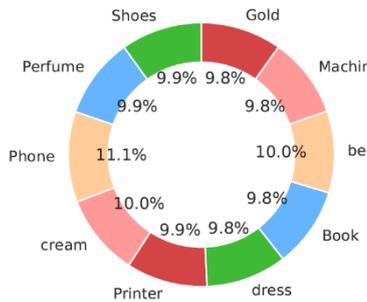
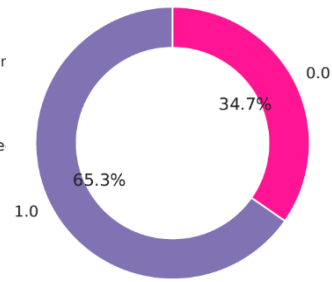
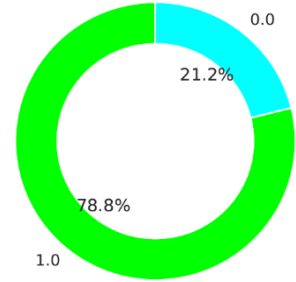

Figure 11: All data domain frequency.

Figure 12: jewelry domain labels frequency.

Figure 13: All domain labels frequency.

### 5- Model evaluation criteria:

There are different evaluation criteria for binary classifications, one of them is the use of a clutter matrix where each record indicates positive or negative. FP (False positive) is the number of negative samples that are predicted positive, TN (True Negative) is the number of negative samples that are classified as negative, and FN (False Negative) is the number of positive samples that are categorized as negative. Here are the criteria for defining these criteria as follows:

$$Precision = \frac{TP}{TP + FP}$$

$$Recall = \frac{TP}{TP + FN}$$

$$Fscore = 2 * \frac{Precision + recall}{Precision * recall}$$

$$Accuracy = \frac{TP + FN}{TP + FP + FN + TN}$$

### 6- Experimental results:
### 6.1 Persian dataset:

In the first step, to prove that WcapsuleE can be useful for multi-domain sentiment analysis, we divided the DigiKala dataset into training and testing data. 80% of the data was used for model training and 20% for model testing. We also test various basic models on these data. The table below summarizes the results obtained by these approaches. As you can see, the proposed approach has achieved higher accuracy than other methods on both emotion classification and domain detection tasks. This correlation between the basic models increases the motivation to deploy the WcapsuleE model to solve the problem of multi-domain sentiment analysis.

Also, the WcapsuleE approach, with the cost function, achieved much higher results. Compared to the WcapsuleE approach, this approach was able to achieve 0.0162, 0.0113 and 0.0218 improvements in accuracy, precision and recall, respectively. This approach also outperformed the WcapsuleE approach in domain classification with 0.0192 improvement in accuracy.

**Table 3: The results of comparing the models on the dataset with 80% training data and 20% testing**.

| Model | Polarity detection | | | | | |
|---|---|---|---|---|---|---|
| | Accuracy | | Precision | | Recall | |
| | Train | Test | Train | Test | Train | Test |
| CNN-Multi Channel(35) | 0.8704 | 0.8775 | 0.8834 | 0.8627 | 0.9331 | 0.9207 |
| Character level CNN (36) | 0.9002 | 0.8823 | 0.9220 | 0.9212 | 0.9312 | 0.9178 |
| NeuroSent(10) | 0.9160 | 0.9183 | 0.9290 | 0.9234 | 0.9222 | 0.9118 |
| Bi-GRUCapsule (37) | 0.9423 | 0.9345 | 0.9231 | 0.9347 | 0.9432 | 0.9336 |
| **WcapsuleE** | **0.9565** | **0.9489** | **0.9706** | **0.9535** | **0.9744** | **0.9524** |
| **WcapsuleE+ Cost sensitivity** | **0.9744** | **0.9699** | **0.9802** | **0.9720** | **0.9832** | **0.9680** |

**Table 4: Results obtained from different approaches for domain detection**.

| Model | Domain identification | | |
|---|---|---|---|
| | Accuracy | Precision | Recall |
| CNN-Multi Channel(35) | 0.6958 | 0.8582 | 0.7255 |
| Character level CNN (36) | 0.7043 | 0.8653 | 0.7589 |
| NeuroSent(10) | 0.7377 | 0.8982 | 0.8290 |
| Bi-GRUCapsule (37) | 0.7809 | 0.8909 | 0.8554 |
| **WcapsuleE** | **0.8020** | **0.9223** | **0.8829** |
| **WcapsuleE+ Cost sensitivity** | **0.8212** | **0.9544** | **0.9212** |

Cross-validation analysis of the WcapsuleE approach with fold=5 shows the effectiveness of this approach. Table 8 shows the results obtained with the proposed approach in DigiKala. Compared to other methods such as Bi-GRUCapsule. NeuroSent, Char-CNN, and multi-channel CNN, this approach has obtained acceptable results. The multi-channel CNN model has lower average accuracy than other proposed methods. This model has an accuracy of less than 0.96% in all domains, and its average accuracy is 0.8347.

In contrast, the CNN character level model has achieved higher accuracy than multi-channel in most domains. The NeuroSent, an LSTM-based model, has achieved relatively good accuracy in some domains. However, this model achieved lower average accuracy than Bi-GRUCapsule and WcapsuleE. The Bi-GRUCapsule model has similar functionality to WcapsuleE in many areas. The WcapsuleE model is more accurate than other approaches in most domains. The model achieved an average accuracy of 0.9144, which is 0.07% better than the weakest approach and 0.03% better than the Bi-GRUCapsule.

| Domain | NeuroSent | Multi-channel CNN | Character level CNN | Bi-GRUCapsule | WcapsuleE | WcapsuleE+ Cost sensitivity |
|---|---|---|---|---|---|---|
| Shoes | **0.8717** | 0.8232 | 0.8514 | 0.8692 | 0.9134 | **0.9214** |
| perfume | 0.8837 | 0.8122 | 0.8420 | 0.8864 | 0.8917 | **0.9070** |
| phone | 0.8718 | 0.8105 | 0.8097 | 0.8693 | **0.9232** | **0.9120** |
| cold cream | 0.8650 | 0.8402 | 0.8728 | 0.8652 | **0.9291** | **0.9167** |
| printer | 0.8866 | 0.8054 | 0.8093 | 0.8915 | 0.9272 | **0.9321** |
| dress | 0.8996 | 0.8434 | 0.8714 | 0.8762 | 0.9075 | **0.9270** |
| Book | 0.8941 | 0.8455 | 0.8700 | 0.8926 | 0.9094 | **0.9119** |
| Bed | 0.8911 | 0.8612 | 0.8543 | 0.8911 | **0.9114** | **0.9001** |
| Shaving machine | 0.9086 | 0.8491 | 0.8704 | 0.8806 | 0.9125 | **0.9222** |
| jewelry | 0.8890 | 0.8567 | 0.8589 | 0.8828 | 0.9193 | **0.9231** |
| Average | 0.8861 | 0.8347 | 0.8510 | 0.8804 | **0.9144** | **0.9173** |

**7- Error analysis and future works**:

Figures 17-26 show the TP, FP, FN, and TN rates obtained by the proposed WcapsuleE approach on the test data. As shown in Figure 26-17, the Shoes and Perfume domains have the highest FP value, and Perfume and dress domains have the highest TN values.

For future work to improve the method proposed in this article, the following methods can be suggested:

1. Detection of sarcasm with a new algorithm
2. Using more preprocessing methods to reduce noise in the collected comments, such as replacing some irregular forms of words with their correct forms
3. In general, one of the problems of using pre-trained word embedding methods is that the calculated word vectors do not contain emotional information. In [56], the authors proposed an Improved Word Vector (IWV) to solve this problem. We hope to have more improvements in our results by combining this algorithm with WcapsuleE.

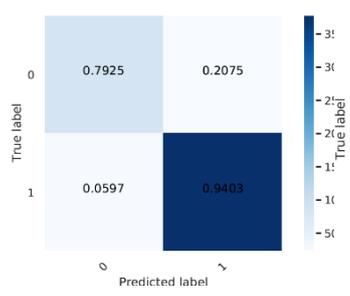
**Figure 14:book**

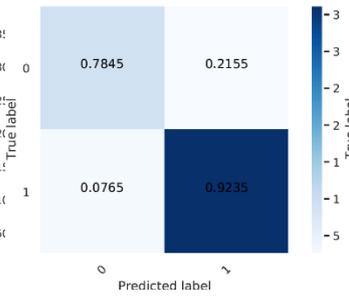
**Figure 15:Perfume**

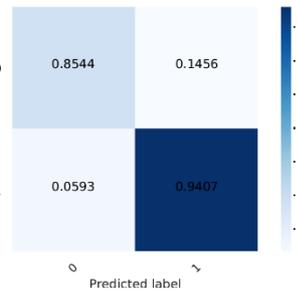
**Figure 16:Phone**

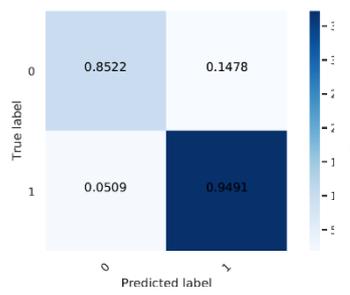
**Figure 17:printer**

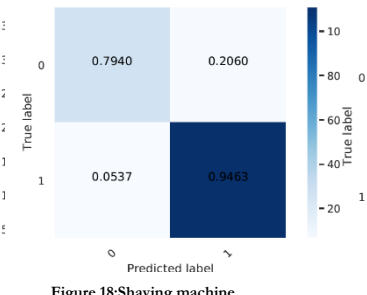
**Figure 18:Shaving machine**

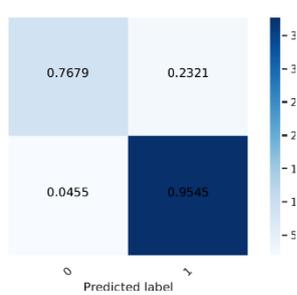
**Figure 19:Shoes**

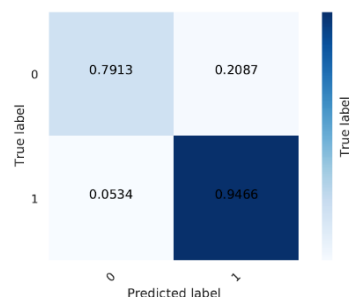
**Figure 20: bed**

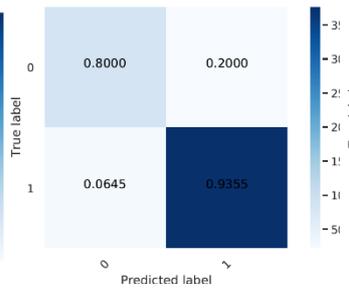
**Figure 21:dress**

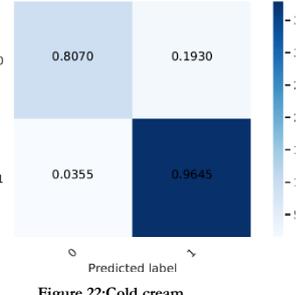
**Figure 22:Cold cream**

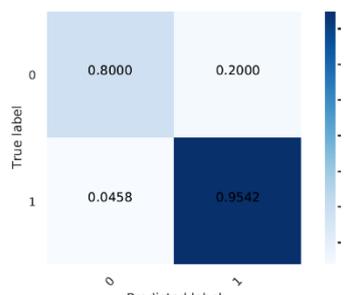
**Figure 23: Jewellery**

**Arabic dataset**:

In this dataset, we tried to collect the opinions of users who have purchased from Amazon. For this purpose, four different domains were collected: clothing, books, jewelry, software, and perfumes. The collection code was adjusted under Python 3.6. The comments with more than four stars give a rate of 1, and the comments with two and less than two stars share a zero rate.

The proposed and other approaches' results are presented on the Arabic dataset. The CNN-Multi Channel approach (35) achieved an accuracy of 0.8024 in this data set for sentiment classification, which is the lowest value obtained in this data set. Character level CNN approach (36) performed better than CNN-Multi Channel and achieved 0.8482 accuracies, and also the improvement of this approach in Precision and Recall criteria is comparable. NeuroSent (10) has performed relatively better than the two examined approaches, but its progress was not as significant as its Character level CNN. Bi-GRUCapsule was an approach that yielded very poor results. This approach achieved a deftness of 0.8235, which was the worst result after the CNN-Multi Channel.

| Model | Polarity classification | | |
|---|---|---|---|
| | Accuracy | Precision | Recall |
| CNN-Multi Channel(35) | 0.8024 | 0.8829 | 0.8718 |
| Character level CNN (36) | 0.8482 | 0.9104 | 0.9029 |
| NeuroSent(10) | 0.8525 | 0.9201 | 0.9001 |
| Bi-GRUCapsule (37) | 0.8235 | 0.9111 | 0.9213 |
| **WcapsuleE** | **0.8607** | **0.9129** | **0.9199** |
| **WcapsuleE+ Cost sensitivity** | **0.8698** | **0.9217** | **0.9311** |

The two proposed approaches WcapsuleE and WcapsuleE+ cost sensitivity achieved the highest accuracy among the approaches on this data set. These two approaches attained an accuracy of 0.8607 and 0.8698, respectively.

In domain classification, most of the approaches achieved a high accuracy of 0.93. The proposed method in this classification attained an accuracy of 0.9609 in the case without the sensitivity function and 0.9695 in the case with the sensitivity function.

| Model | Domain classification | | |
|---|---|---|---|
| | Accuracy | Precision | Recall |
| CNN-Multi Channel(35) | 0.9321 | 0.9820 | 0.9918 |
| Character level CNN (36) | 0.9482 | 0.9964 | 0.9929 |
| NeuroSent(10) | 0.9521 | 0.9932 | 0.9901 |
| Bi-GRUCapsule (37) | 0.9331 | 0.9922 | 0.9913 |
| **WcapsuleE** | **0.9609** | **0.9913** | **0.9999** |
| **WcapsuleE+ Cost sensitivity** | 0.9695 | 0.9915 | 0.9911 |

Three basic models, including Bi-GRUCapsule, WcapsuleE, and WcapsuleE+ Cost sensitivity, are used in our analysis. For sensitivity analysis, the effectiveness of two critical parameters has been studied on these basic models: batch size and dimensions of the hidden layer.

• **Batch size:** The batch size represents the number of data samples to be transmitted over the network. The larger this parameter is, the more memory the network consumes, and the smaller its value, the longer the network training time. For this purpose, different values of 8, 16, 32, 64, 128, and 256 have been evaluated for this parameter. Figures 4, 6, and 8 show different batch size values for the base models. Based on this test, 128, 8, and 128 have been selected as the batch sizes for the basic models (GRUCapsule, WcapsuleE, and WcapsuleE+ Cost sensitivity).

• **Hidden layer dimension:** Deciding on the number of neurons in hidden layers is essential to the overall architecture of neural networks. Using very few neurons in hidden layers leads to the problem of the misfit. On the contrary, using many neurons in hidden layers can also cause several problems, such as overfitting. Therefore, a balanced number of neurons should be used in the layers. Thus, the number of different neurons, including 8, 16, 32, 64, and 128, have been investigated in our experiments. Figures 5, 7, and 9 show different values of hidden layer dimensions for basic models. Based on this test, 64, 64, and 64 have been selected for the dimensions of the hidden layer of the three basic models (GRUCapsule, WcapsuleE, and WcapsuleE+ Cost sensitivity).

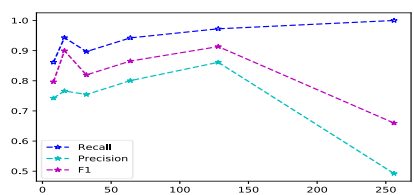
Figure 4: Batch size and evaluation criteria

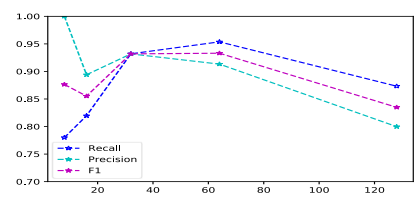
Figure 5: Hidden layer size and evaluation criteria

Effect of batch size and hidden layer size in GRUCapsule model

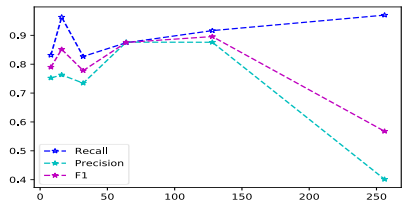
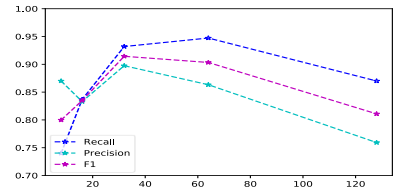

Figure 6: Batch size and evaluation criteria

Figure 7: Hidden layer size and evaluation criteria

Effect of batch size and hidden layer size in WcapsuleE model

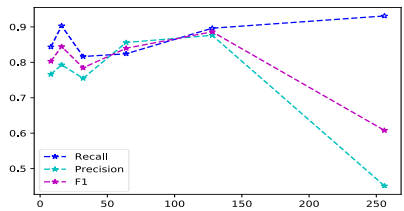
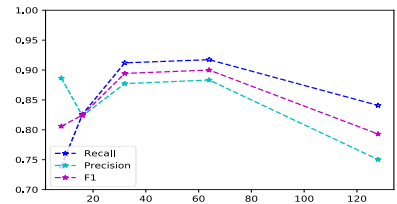

Figure 8: Batch size and evaluation criteria

Figure 9: Hidden layer size and evaluation criteria

Effect of batch size and hidden layer size in WcapsuleE+ cost sensitivity model

### 8- Conclusion and future works:

In this paper, we proposed the WcapsuleE method based on two-way GRU and CapsuleNet for multi-domain sentiment analysis using the DBD domain dependency measure to infer the polarity of documents. This approach included embedded words, bidirectional GRU, encapsulation, classification layer, and DBD criterion. The efficiency of the proposed method was evaluated using Digikala and Amazon Arabic data, and the results show the success of the proposed method compared to the relevant advanced systems. On the other hand, the proposed approach has been able to consider the effect of negative words due to maintaining the features as a set of vectors.